\documentclass{egpubl}
\usepackage{eg2023}
\usepackage[T1]{fontenc}
\usepackage{dfadobe}  

\usepackage{cite}  % comment out for biblatex with backend=biber
\BibtexOrBiblatex

\usepackage{times}
\usepackage{epsfig}
\usepackage{graphicx}
\usepackage{amsmath}
\usepackage{amssymb}
\usepackage{booktabs}
\usepackage{pifont}
\usepackage{multirow}
\usepackage{makecell}
\usepackage{xcolor}
\usepackage{ifsym}
\usepackage{etoolbox}
\usepackage{xspace}
\usepackage{dcolumn}

\usepackage{soul}
\usepackage{amsmath}
\usepackage{amssymb}
\usepackage{microtype}
\usepackage{wrapfig}
\usepackage{fancyhdr}

\def\figurePath{figures/}
\def\myfigure#1#2{\begin{figure}[htb]\centering\includegraphics*[width = \linewidth]{\figurePath#1}\centering\caption{#2}\label{fig:#1}\vspace{-0.2cm}\end{figure}}

\def\mycfigure#1#2{\begin{figure*}[t]\centering\includegraphics*[clip, width = \linewidth]{\figurePath#1}\centering\caption{#2}\label{fig:#1}\vspace{-0.2cm}\end{figure*}}

\newcommand{\refSec}[1]{Sec.~\ref{sec:#1}}
\newcommand{\refFig}[1]{Fig.~\ref{fig:#1}}
\newcommand{\refEq}[1]{Eq.~\ref{eq:#1}}
\newcommand{\refTbl}[1]{Tbl.~\ref{tbl:#1}}

\soulregister\ref7
\soulregister\cite7
\soulregister\refFig7
\soulregister\refSec7
\soulregister\ref7
\soulregister\pageref7
\soulregister\shortcite7
\soulregister\eg0
\soulregister\ie0

\DeclareGraphicsExtensions{.png,.jpg,.pdf,.ai,.psd}
\newcommand{\mysection}[2]{\section{#1}\label{sec:#2}}
\newcommand{\mysubsection}[2]{\subsection{#1}\label{sec:#2}}

\definecolor{colorA}{HTML}{777777}
\definecolor{colorB}{HTML}{4285f4}
\definecolor{colorC}{HTML}{ea4335}
\definecolor{colorD}{HTML}{fbbc04}
\definecolor{colorE}{HTML}{34a853}
\definecolor{colorF}{HTML}{ff6d01}
\definecolor{colorG}{HTML}{46bdc6}
\definecolor{colorH}{HTML}{000000}

\newcommand{\colorIcon}[2]{\textcolor{color#1}{\csname icon#2\endcsname}}

\newcommand{\nameAndIcon}[1]{``#1'' ({\csname icon#1\endcsname})}
\newcommand{\nameAndIconA}[1]{(``#1'', {\csname icon#1\endcsname})}

\usepackage[pagebackref=true,breaklinks=true,colorlinks,bookmarks=false]{hyperref}

\newcommand{\myparagraph}[1]{\vspace{0.15cm} \noindent\textbf{#1}\ \ }
\newcommand{\mymath}[2]{\newcommand{#1}{\TextOrMath{$#2$\xspace}{#2}}}

\mymath{\previousFrame}{\mathtt{{SL}_{0}}}
\mymath{\currentFrame}{\mathtt{{SL}_{1}}}
\mymath{\midFrame}{\mathtt{\hat{I}_{t}}}
\mymath{\midFrameLevel}{\mathtt{\hat{I}^{l}_{t}}}

\mymath{\midFrameGT}{\mathtt{I_{t}}}
\mymath{\midFrameGTLevel}{\mathtt{I^{l}_{t}}}

\mymath{\denoiseNet}{\pmb{\mathtt{denoise}}}

\mymath{\previousHigh}{\mathtt{H^{e}_{0}}}
\mymath{\previousHighBlurry}{\mathtt{H^{b}_{0}}}
\mymath{\previousLow}{\mathtt{L^{e}_{0}}}
\mymath{\currentHigh}{\mathtt{H^{e}_{1}}}
\mymath{\currentHighBlurry}{\mathtt{H^{b}_{1}}}
\mymath{\currentLow}{\mathtt{L^{e}_{1}}}
\mymath{\currentHdr}{\mathtt{Y^{e}_{1}}}
\mymath{\currentHdrStart}{\mathtt{Y^{s}_{1}}}
\mymath{\previousHdrStart}{\mathtt{Y^{s}_{0}}}
\mymath{\previousHdr}{\mathtt{Y^{e}_{0}}}
\mymath{\gtBlurryHdr}{\mathtt{Y^{b_{i}}_{t}}}
\mymath{\gtHdrTi}{\mathtt{Y_{t}}}
\mymath{\currentSpatialFlow}{\mathtt{F_{1}}}
\mymath{\previousSpatialFlow}{\mathtt{F_{0}}}
\mymath{\currentSpatialFlowFirst}{\mathtt{f^{b_1}_{0 \rightarrow -1}}}
\mymath{\currentSpatialFlowSecond}{\mathtt{f^{b_2}_{0 \rightarrow -1}}}
\mymath{\currentSpatialFlowThird}{\mathtt{f^{b_3}_{0 \rightarrow -1}}}
\mymath{\previousSpatialFlowFirst}{\mathtt{f^{b_1}_{-1 \rightarrow -2}}}
\mymath{\previousSpatialFlowSecond}{\mathtt{f^{b_2}_{-1 \rightarrow -2}}}
\mymath{\previousSpatialFlowThird}{\mathtt{f^{b_3}_{-1 \rightarrow -2}}}
\mymath{\currentSpatialFlowT}{\mathtt{f^{b_t}_{0 \rightarrow -1}}}
\mymath{\warpedCurrentToBlurryT}{\mathtt{Y^{b}_{0 \rightarrow t}}}
\mymath{\warpedCurrentToBlurryTi}{\mathtt{Y^{b_{i}}_{0 \rightarrow t}}}
\mymath{\warpedCurrentToBlurryThree}{\mathtt{Y^{b_3}_{0 \rightarrow t}}}
\mymath{\warpedCurrentToT}{\mathtt{Y_{0 \rightarrow t}}}
\mymath{\warpedCurrentToTi}{\mathtt{Y^{i}_{0 \rightarrow t}}}
\mymath{\flowTemporalForward}{\mathtt{f^{t}_{0 \rightarrow -1}}}
\mymath{\flowTemporalBackward}{\mathtt{f^{t}_{-1 \rightarrow 0}}}
\mymath{\flowTemporalBackwardSplit}{\mathtt{f^{t}_{-1 \rightarrow t}}}
\mymath{\flowTemporalForwardSplit}{\mathtt{f^{t}_{0 \rightarrow t}}}

\mymath{\exptime}{\mathtt{t_{e}}}
\mymath{\readtime}{\mathtt{t_{R}}}

\mymath{\warp}{\mathtt{warp}}
\mymath{\loss}{\mathtt{L}_}

\mathchardef\mhyphen="2D % Define a "math hyphen"

\mymath{\denseNet}{\mathtt{D\mhyphen Net}}

\mymath{\interlaceprev}{\mathtt{{M}_{0}}}
\mymath{\interlacecurr}{\mathtt{{M}_{1}}}
\mymath{\interlacecurrcurr}{\mathtt{{M}_{2}}}
\mymath{\interlace}{\mathtt{{M}_{i}}}

\mymath{\longBlurry}{\mathtt{{\hat{L}}_{i}}} 
\mymath{\longBlurryFeat}{\mathtt{{{U}}^{l}_{i}}}
\mymath{\longBlurryFeatPrev}{\mathtt{{{U}}^{l}_{0}}}
\mymath{\longBlurryFeatCurr}{\mathtt{{{U}}^{l}_{1}}}

\mymath{\longBlurryGT}{\mathtt{{L}_{i}}}
\mymath{\longBlurryPrev}{\mathtt{\hat{L}_{0}}}
\mymath{\longBlurryCurr}{\mathtt{\hat{L}_{1}}}

\mymath{\flowPrevCurrEndLevel}{\mathtt{{F}^{l}_{e0\rightarrow e1}}}
\mymath{\flowPrevCurrStartLevel}{\mathtt{{F}^{l}_{e0\rightarrow s1}}}
\mymath{\flowCurrPrevEndLevel}{\mathtt{{F}^{l}_{e1\rightarrow e0}}}
\mymath{\flowCurrPrevStartLevel}{\mathtt{{F}^{l}_{e1\rightarrow s0}}}

\mymath{\HDRsharp}{\mathtt{{\hat{I}}_{ei}}}
\mymath{\HDRsharpStart}{\mathtt{{\hat{I}}_{si}}}
\mymath{\HDRsharpGT}{\mathtt{{I}_{ei}}}

\mymath{\HDRsharpFeat}{\mathtt{{V}^{l}_{i}}}
\mymath{\HDRsharpFeatPrev}{\mathtt{{V}^{l}_{0}}}
\mymath{\HDRsharpFeatCurr}{\mathtt{{V}^{l}_{1}}}

\mymath{\sharpPrevStart}{\mathtt{{\hat{I}}_{s0}}}
\mymath{\sharpPrevEnd}{\mathtt{{\hat{I}}_{e0}}}
\mymath{\sharpCurrStart}{\mathtt{{\hat{I}}_{s1}}}
\mymath{\sharpCurrEnd}{\mathtt{{\hat{I}}_{e1}}}

\mymath{\sharpPrevStartLevel}{\mathtt{{\hat{I}^{l}}_{s0}}}
\mymath{\sharpCurrStartLevel}{\mathtt{{\hat{I}^{l}}_{s1}}}
\mymath{\sharpPrevEndLevel}{\mathtt{{\hat{I}}^{l}_{e0}}}
\mymath{\sharpCurrEndLevel}{\mathtt{{\hat{I}}^{l}_{e1}}}

\mymath{\sharpPrevEndGT}{\mathtt{{I}_{ei}}}
\mymath{\sharpPrevStartGT}{\mathtt{{I}_{si}}}

\mymath{\sharpPrevEndGTLevel}{\mathtt{{I}^{l}_{ei}}}
\mymath{\sharpPrevStartGTLevel}{\mathtt{{I}^{l}_{si}}}

\mymath{\warpPrev}{\mathtt{{\hat{I}}_{0\rightarrow t}}}
\mymath{\warpCurr}{\mathtt{{\hat{I}}_{1\rightarrow t}}}

\mymath{\warpPrevLevel}{\mathtt{{\hat{I}}^{l}_{0\rightarrow t}}}
\mymath{\warpCurrLevel}{\mathtt{{\hat{I}}^{l}_{1\rightarrow t}}}

\mymath{\flowInside}{\mathtt{{F}_{ei \rightarrow si}}}
\mymath{\flowInsideReverse}{\mathtt{{F}_{si \rightarrow ei}}}

\mymath{\flowInsideLevel}{\mathtt{{F}^{l}_{ei \rightarrow si}}}
\mymath{\flowInsideLevelprev}{\mathtt{{F}^{l+1}_{ei \rightarrow si}}}

\mymath{\flowPrev}{\mathtt{{F}_{e0 \rightarrow s0}}}
\mymath{\flowCurr}{\mathtt{{F}_{e1 \rightarrow s1}}}
\mymath{\flowCurrReverse}{\mathtt{{F}_{s1 \rightarrow e1}}}

\mymath{\flowPrevLevel}{\mathtt{{F}^{l}_{e0 \rightarrow s0}}}
\mymath{\flowCurrLevel}{\mathtt{{F}^{l}_{e1 \rightarrow s1}}}

\mymath{\alphaLevel}{\mathtt{\alpha^{l}}}
\mymath{\alphaLevelPrev}{\mathtt{\alpha^{l+1}}}

\mymath{\flowPrevCurrEnd}{\mathtt{{F}_{e0\rightarrow e1}}}
\mymath{\flowPrevCurrStart}{\mathtt{{F}_{e0\rightarrow s1}}}

\mymath{\flowCurrPrevEnd}{\mathtt{{F}_{e1\rightarrow e0}}}
\mymath{\flowCurrPrevStart}{\mathtt{{F}_{e1\rightarrow s0}}}

\mymath{\backflowprev}{\mathtt{{F}_{t\rightarrow 0}}}
\mymath{\backflowcurr}{\mathtt{{F}_{t\rightarrow 1}}}
\mymath{\forwardprev}{\mathtt{{F}_{0 \rightarrow t}}}
\mymath{\forwardcurr}{\mathtt{{F}_{1\rightarrow t}}}

\mymath{\backflowprevLevel}{\mathtt{{F}^{l}_{t\rightarrow 0}}}
\mymath{\backflowcurrLevel}{\mathtt{{F}^{l}_{t\rightarrow 1}}}
\mymath{\forwardprevLevel}{\mathtt{{F}^{l}_{0 \rightarrow t}}}
\mymath{\forwardcurrLevel}{\mathtt{{F}^{l}_{1\rightarrow t}}}

\mymath{\quadmodelprev}{\mathtt{\dfrac{1}{2} {a}_{0} \times t^{2}+ {v}_{0} \times t}}
\mymath{\quadmodelcurr}{\mathtt{\dfrac{1}{2} {a}_{1} \times t^{2}+ {v}_{1} \times t}}

\mymath{\makehdr}{\pmb{\mathtt{MakeHDR}}}
\mymath{\blurtoflow}{\pmb{\mathtt{Blur2Flow}}}
\mymath{\fitQuad}{\pmb{\mathtt{FitQuad}}}
\mymath{\reverse}{\pmb{\mathtt{Reverse}}}
\mymath{\blend}{\pmb{\mathtt{Blend}}}
\mymath{\suffixE}{\pmb{\mathtt{e}}}
\mymath{\suffixS}{\pmb{\mathtt{s}}}

\mymath{\MixedFeature}{\mathtt{{W}^{l}}}
\mymath{\MixedFeaturePrev}{\mathtt{{W}^{l-1}}}

\definecolor{fixedcolor}{rgb}{1,.95,.4}

\definecolor{red}{rgb}{0.7,0,0}
\title{Video frame interpolation for high dynamic range sequences captured with dual-exposure sensors}
\author[\c{C}o\u{g}alan et al.]
{\parbox{\textwidth}{\centering U\u{g}ur \c{C}o\u{g}alan, Mojtaba Bemana, Hans-Peter Seidel and Karol Myszkowski
	}
	\\
	{\parbox{\textwidth}{\centering Max-Planck-Institut f\" ur Informatik, Germany
		}
	}
}
\begin{document}

\teaser{
 \vspace{-1.0cm}
 \includegraphics[width=\linewidth]{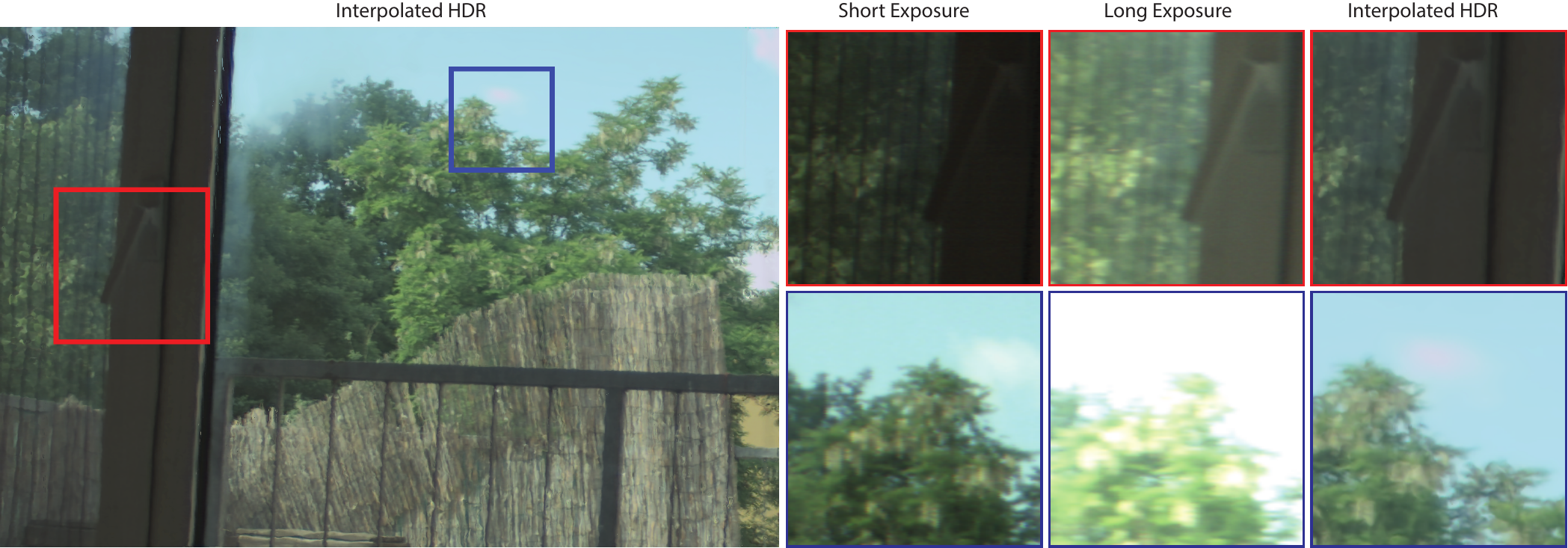}
 \centering
  \caption{
  %HDR video reconstruction of a high-contrast scene (left) using different methods, including ours (right, columns), and different sensors (rows 1 and 2 vs. rows 3 and 4).
  We propose a method for high dynamic range (HDR) video frame interpolation (VFI) for dual-exposure sensors that gain on popularity due to their use in recent smartphones.
  The first column shows interpolated HDR frame, while the insets focus on the dark and bright scene details. Note that the short and long exposures, as captured by the sensor (middle columns), are shifted with respect to the interpolated HDR frame (right column). The dark region (the upper row) requires a long exposure duration and features significant motion blur due to camera motion.
  Our method employs temporally continuous information on the scene motion that is encoded in motion blur to improve the VFI quality.
  At the same time, the short exposure avoids pixel saturation in the sky region (the bottom row) and enables its reconstruction in the interpolated HDR frame. 
  }
\label{fig:teaser}
}

% uncomment for using teaser
% \teaser{
%  \includegraphics[width=\linewidth]{eg_new}
%  \centering
%   \caption{New EG Logo}
% \label{fig:teaser}
%}

\maketitle
%-------------------------------------------------------------------------
\begin{abstract}
% Video frame interpolation (VFI) enables many important applications that might involve the temporal domain, such as slow motion playback, or the spatial domain, such as stop motion sequences. 
% We are focusing on the former task, where one of the key challenges is handling high dynamic range (HDR) scenes in the presence of complex motion.
%Video frame interpolation (VFI) enables many important applications \hl{that involves the manipulation of time, such as slow motion playback, or smoothing stop motion sequences.
Video frame interpolation (VFI) enables many important applications such as slow motion playback and frame rate conversion. %or smoothing stop motion sequences. 
However, one major challenge in using VFI is accurately handling high dynamic range (HDR) scenes with complex motion.
To this end, we explore the possible advantages of dual-exposure sensors that readily provide sharp short and blurry long exposures that are spatially registered and whose ends are temporally aligned.
This way, motion blur registers temporally continuous information on the scene motion that, combined with the sharp reference, enables more precise motion sampling within a single camera shot. 
We demonstrate that this facilitates a more complex motion reconstruction in the VFI task, as well as HDR frame reconstruction that so far has been considered only for the originally captured frames, not in-between interpolated frames.
We design a neural network trained in these tasks that clearly outperforms existing solutions.
We also propose a metric for scene motion complexity that provides important insights into the performance of VFI methods at test time.

%-------------------------------------------------------------------------
%  ACM CCS 1998
%  (see https://www.acm.org/publications/computing-classification-system/1998)
% \begin{classification} % according to https://www.acm.org/publications/computing-classification-system/1998
% \CCScat{Computer Graphics}{I.3.3}{Picture/Image Generation}{Line and curve generation}
% \end{classification}
%-------------------------------------------------------------------------
%  ACM CCS 2012 (see https://www.acm.org/publications/c 

%The tool at \url{http://dl.acm.org/ccs.cfm} can be used to generate
% CCS codes.



% \ccsdesc[300]{Computational photography}
% \ccsdesc[300]{High Dynamic Range Imaging}
% \ccsdesc[100]{Video Frame Interpolation}

\end{abstract}  
%-------------------------------------------------------------------------
\mysection{Introduction}{Introduction}
Video frame interpolation (VFI) enables many interesting applications ranging from video compression and framerate up-conversion in TV broadcasting to artistic video effects such as speed ramp in professional cinematography. The performance of VFI methods is largely affected by various factors such as scene lighting conditions, the magnitude and complexity of motion in the scene, the spatial extension of resulting motion blur, the presence of complex occlusions, or thin structures in the scene. Popular VFI methods \cite{jiang2018super,bao2019depth,sim2021xvfi} mostly rely on well-exposed frames in the captured video.
Nevertheless, in the case of high dynamic range (HDR) scenes captured using traditional single-exposure sensors, undesired under- and over-exposure effects might appear. The resultant noise and intensity clamping can adversely affect the quality of VFI as finding the pixel correspondence between the frames becomes more ambiguous. Another major challenge is the large and non-uniform motion in the scene. Although recent methods \cite{reda2022film,sim2021xvfi} have shown progress in handling large motion, they typically heavily rely on the motion linearity assumption that might not hold in practice. 
Explicit handling of non-linear motion becomes possible by processing more than two subsequent frames\cite{xu2019quadratic,park2021asymmetric}; however, temporal sampling might still be too low for reliable motion reconstruction. 
Motion blur due to low shutter speed and long exposure times further leads to spatial and temporal loss of image details. For this reason, handling blurry frames is typically treated as a challenge in the VFI task \cite{shen2020blurry,zhang2020video}, while potentially, motion blur encodes continuous temporal information on the magnitude and direction of motion, particularly for large motion.

Programmable sensors with spatially varying exposures greatly expand the dynamic range of contrast in captured video \cite{hajsharif2014hdr,go2019image,choi2017reconstructing,heide2014flexisp,cogalan2022learning,cho2014single}, and become an attractive choice for modern smartphones \cite{quadBayerphone}, e.g.\, Sony's Quad Bayer \cite{quadBayer}, and Samsung's Tetracell/Nonacell \cite{samsungisocell-gn1} technologies.
% Programmable sensors with spatially varying exposures \cite{quadBayer,carey2013100} greatly expand the dynamic range of contrast in captured scenes \cite{hajsharif2014hdr,go2019image,choi2017reconstructing,heide2014flexisp,cogalan2022learning,cho2014single}.
% \added{An HDR image reconstructed using such sensors would not only provide a greater illumination range than general low dynamic range (LDR) images but also the better clarity of detail in highlights and dark regions can potentially improve the performance of many vision and graphics tasks such as object detection, depth estimation, scene segmentation, and etc. Novel sensor designs, offering great flexibility of in-pixel processing, have been successfully used in video compressive imaging \cite{martel2020neural,iliadis2020deepbinarymask}, depth from defocus \cite{martel2017high}, feature classification \cite{chen2017feature}, HDR imaging \cite{martel2020neural} and motion deblurring \cite{nguyen2022learning}. The functionality of novel sensor designs in the context of HDR imaging and its possible applications have been comprehensively discussed in the survey \cite{wang2021deep}. }
In this work, we explore such sensor capabilities toward improving the motion estimation accuracy in VFI.
In particular, we consider a dual-exposure sensor that captures short and long exposures for spatially interleaved pixel columns in a single shot \cite{cmosis2020}.
Importantly, while the exposure duration differs, the exposure completion is temporally aligned, which enables recovering two temporal samples of the scene motion that are perfectly spatially registered at the sensor. 
We show that such an increased temporal sampling rate substantially improves the accuracy of complex motion interpolation, as motion non-linearity can readily be reconstructed for two subsequent frames.
Furthermore, the short exposure typically leads to a sharp image, while the long exposure results in substantial motion blur that provides additional insights into the motion direction and magnitude (\refFig{teaser}).
This is of particular importance in dark scene regions, where the short exposure might be strongly underexposed and noisy, and the long exposure becomes the only reliable measurement of scene motion.
As in other works, we employ a multi-exposure technique to reconstruct HDR video frames, but for the first time, we simultaneously perform VFI that can handle complex, non-linear motion in the scene.
We train an end-to-end convolutional network to achieve those goals.
We also propose a metric of motion non-linearity that allows us to analyze the existing high-speed videos and measure the performance of VFI methods as a function of motion complexity. % and derive a balanced training dataset that greatly improves our network generality in complex motion handling. 

The key contributions of our work are:
\begin{itemize} 
\item We propose a compact machine learning solution for VFI that can handle HDR content and complex non-uniform motion, enabled by deriving two temporal samples of the scene motion for each frame by joint processing of short and long exposures as captured using a dual-exposure sensor. 
\item We adopt a PWC-Net architecture to estimate the motion flow from motion blur in the long exposure that, in our setup, is uniquely supported by sharp image content in the short exposure. Spatial registration of both exposures and temporal alignment of their ends greatly improves the motion flow accuracy.
\item We develop a metric of motion complexity that provides interesting insights into existing datasets used in the training of VFI methods and enables us to evaluate the performance of those methods for different levels of motion non-linearity.

%\item We develop a strategy of balancing training data in terms of motion non-linearity that greatly improves our network ability to a meaningful motion interpolation.
\end{itemize}

In the following section, we discuss previous work, and in \refSec{sec_method}, we present our VFI method for HDR sequences. In \refSec{sec_NonlinearityMetric} we introduce our metric of scene motion uniformity that enables meaningful comparison of existing VFI methods while
\refSec{impl} provides implementation details of our network. \refSec{Results_} contrasts our technique with existing works in a performance comparison and reports an outcome of ablation studies. Finally, we conclude this work in \refSec{Conclusion_}.

%  Slow motion playback in sport matches and interpolation between the keyframes in the stop motion videos are among the interesting applications for the video frame interpolation method.

\mysection{Previous work}{PrevWork}

In this section, we discuss existing VFI methods dealing with sharp input video (\refSec{rw_SharpInterp}), considering either a uniform or non-uniform motion assumption. 
We focus on the problems of recovering motion from the blur (\refSec{rw_motion_from_blur}), joint deblurring and VFI (\refSec{rw_DeblurInterp}), and HDR video reconstruction (\refSec{rw_HDRvideo}) that are central to this work. 
We refer the reader to recent surveys where more complete treatments of deep VFI \cite{parihar2021comprehensive} and HDR video \cite{wang2021deep} solutions are presented.

\mysubsection{Sharp video frame interpolation}{rw_SharpInterp}

%Recent video interpolation methods that deal with sharp video frames either have a uniform motion model assumption \cite{niklaus2017video,jiang2018super,bao2019depth,liu2019deep,reda2019unsupervised,park2020bmbc,choi2020channel,niklaus2020softmax,lee2020adacof,sim2021xvfi,reda2022film} or explicitly non-uniform assumption \cite{xu2019quadratic,chi2020all,Choi2021HQFrameInterpolation,park2021asymmetric}.

A vast majority of existing VFI techniques assume that the motion in the input video is uniform, but there are also methods explicitly designed without this assumption.

\myparagraph{Uniform motion}  SepConv \cite{niklaus2017video} merges flow estimation and frame warping into a single convolution step. They predict spatially-varying 1D kernels and convolve with them input frames to interpolate new frames. SuperSlowMo \cite{jiang2018super} uses bi-directional flows and an occlusion map to synthesize intermediate frames at arbitrary times. DAIN \cite{bao2019depth} utilizes additional interpolation kernels and depth maps for blending the input frames. A cycle consistency loss is introduced to learn frame interpolation with fewer training pairs \cite{liu2019deep}, or without any supervision, \cite{reda2019unsupervised}. BMBC \cite{park2020bmbc} warps the input frames with a proposed bilateral motion model and combines them using learned dynamic blending filters. CAIN \cite{choi2020channel} uses a channel attention module to interpolate video frames without the need for estimation of motion. SoftSplat \cite{niklaus2020softmax} proposes differentiable forward warping via softmax splatting and shows its benefits for VFI. AdaCoF \cite{lee2020adacof} proposes a warping module in which a target pixel can refer to not only one but many pixels at any location in the reference. XVFI \cite{sim2021xvfi} presents a high-speed (1000fps) video dataset and proposes a multi-scale recursive approach to handle large motion in the scene. Recently, FILM \cite{reda2022film} has introduced a unified framework that achieves superior results for large and complex motions by balancing the motion range distribution in the training dataset. 
For all methods discussed here, a combination of large and strongly non-uniform motion might lead to highly objectionable artifacts. 
%As these approaches explicitly or implicitly assume the motion in the input video is linear, they often fail in the case of non-uniform motion in the scene. 

\myparagraph{Non-uniform motion} QVI \cite{xu2019quadratic} is one of the first video interpolation methods to model curvilinear motion with the quadratic equation using four temporal frames. Chi et al. \cite{chi2020all} extend QVI by introducing an additional cubic term that accounts for the change in acceleration. ABME \cite{park2021asymmetric} handles the non-uniform motion in the scene by extending the BMBC \cite{park2020bmbc} for asymmetric bilateral motion between input frames. In all those methods, more than two consecutive frames are required to capture the motion non-uniformity that, for large and complex motions might be challenging, both because of temporal sampling deficits as well as overall reduced flow estimation accuracy. In our approach, we capture two exposures in a single frame that increase the sampling rate twice, and we employ motion blur inherent to the longer exposure as an additional cue to the flow estimation.

\mysubsection{Motion flow reconstruction from motion blur}{rw_motion_from_blur}

A combination of longer exposure times and rapid motion in the scene or camera might lead to visible motion blur that typically is considered degradation and eliminated using the dedicated image and video deblurring solutions. We refer the reader to extensive surveys on this topic \cite{koh2021single,zhang2022deep},
and we focus our discussion on deblurring solutions that explicitly recover intra-frame optical flow from motion blur that we employ in this work. 
Earlier works \cite{rekleitis1995visual,schoueri2009optical} assume global motion models that lead to spatially-invariant deblurring kernels.
More advanced solutions support spatially-varying kernels that are approximated by linear motion  \cite{hyun2014segmentation,dai2008motion}.
Gong et al. \cite{gong2017motion} propose a deep-learning approach to handle heterogeneous blur; however, they simulate motion flows with a set of constrained flow magnitudes and directions to generate the training pairs. Argaw et al. \cite{argaw2021optical} alleviate this issue by deploying available synthetic and real scene blur datasets without any restrictive motion assumptions and estimating a dense optical flow directly from motion blur in the image. However, their estimation may be subject to ambiguity in predicting the correct direction of flow, which is crucial in our case. Beyond restoring latent sharp images, a joint estimate of the 3D shape and motion are feasible, but highly motion-blurred images are required \cite{qiu2019world,rozumnyi2022motion}. While these methods aim to recover the motion flow from blur, we can not apply them right away, as they assume that the input blurry image is mostly well-exposed, while we have a considerable amount of saturated pixels in the blurry long exposure. We deal with this problem using the sharp short exposure that also enables bypassing the task of image deblurring.

%  \added{While these methods are limited to generate only a de-blurred image, a sequence of sharp latent frames can be extracted from a single blurry image. Purohit et al. \cite{purohit2019bringing} generate a video from a motion-blurred image while explicitly taking motion information into account. The method is prone to error propagation, as intermediate sharp sequences are created recursively. Argaw et al. \cite{argaw2021restoration} address this issue by adopting a single-stage frame to restore the underlying video frames from a single motion-blurred image without motion supervision. As there is an ambiguity in temporal ordering of recovered sharp frames, Rengarajan et al. \cite{rengarajan2020photosequencing} adopt a triplet of short-long-short exposures captured by a programmable machine vision camera to restore temporally consistent sequences from the middle long exposure.

% wang2022efficient

\mysubsection{Joint video deblurring and interpolation}{rw_DeblurInterp}
Recent works demonstrate that joint deblurring and frame interpolation greatly improves the resulting VFI quality over an independent treatment of these tasks.
Jin et al. \cite{jin2019learning} adopt a joint optimization scheme to extract sharp keyframes within a frame by processing four consecutive blurry frames and then smoothly interpolating the in-between frame using the extracted keyframes. Shen et al. \cite{shen2020blurry,shen2020video} simultaneously remove the motion blur and interpolate the in-between frames by employing a recurrent pyramid framework to efficiently aggregate the temporal information. Gupta et al. \cite{gupta2020alanet} relax the strong assumption that all the input frames in a captured video are blurry and adapt attention mechanisms to decide on deblurring each frame based on the information from the neighbor frames. While these methods mainly attempt to remove the motion blur in the VFI task, the inherent motion blur, as we discuss in \refSec{rw_motion_from_blur}, can potentially reveal information about the magnitude and direction of the motion, especially in the case of large non-uniform motion. Along these lines Zhang et al. \cite{zhang2020video} propose a VFI solution that is the closest to our work. They first extract two sharp keyframes corresponding to the start and the end of a blurry frame, and then by taking two consecutive frames, they compute the optical flow between the resulting four keyframes. By employing a quadratic motion formulation, they can handle non-uniform motion. 
However, in this approach, the inaccuracy in predicting the keyframes affects the quality of the flow estimation, which in turn is prone to error, especially for large motion, whereas we benefit from the less blurred short exposure in each frame to make the flow estimation more reliable.
This allows us to consider more intra- and inter-frame flows that are independently estimated, and we carry our processing across subsequent stages of our multi-network pipeline using a multiresolution approach.
Also, we uniquely support HDR VFI, so we need to deal with extensive saturation regions in the blurry long exposure.

\mysubsection{HDR video reconstruction}{rw_HDRvideo}
HDR video reconstruction is typically performed using multi-exposure techniques, where subsequent frames with temporally interleaved different exposures are combined, and their dynamic content is aligned, typically using optical flow methods \cite{kalantari2013patch,kalantari2017deep,kalantari2019deep,yan2020deep,chen2021hdr}.
Disparity information can be used for such alignment in stereo cameras or dual-lens systems that are widely used in modern smartphones \cite{lin2009high,chen2019new,chen2020learning,dong2021miehdr}.
To alleviate the need for such alignment, single-shot HDR techniques are developed that rely on specialized dual-ISO/dual-gain sensors that require larger photosites to reduce the photon noise as typically short exposures are captured to avoid highlight clipping \cite{hajsharif2014hdr,go2019image,choi2017reconstructing,cogalan2020deep}. Dedicated hardware solution such as coded sensors using spatially-varying optical mask \cite{serrano2016convolutional,alghamdi2019reconfigurable} can also enable HDR imaging with only a single-shot.
Multi-exposure sensors \cite{cmosis2020,quadBayer,samsungisocell-gn1} that, as we show in this work, are greatly beneficial for HDR VFI, %albeit 
require motion deblurring in longer exposures to reconstruct sharp HDR video  \cite{heide2014flexisp,cho2014single,jiang2021hdr}. 
This task is relatively easy for machine learning solutions \cite{cogalan2022learning}, where recently proposed neural sensors \cite{martel2020neural,nguyen2022learning} can learn spatially varying pixel exposures for efficient motion deblurring.
The scope of all these methods is mainly limited to HDR video reconstruction, and they do not aim for the VFI task. An exception here is the work of Rebecq et al. \cite{rebecq2019high}, where high framerate HDR video is reconstructed using a highly specialized event camera that, in a frameless manner, asynchronously responds to per-pixel brightness changes.

%Our HDR VFI pipeline takes as the input two subsequent frames \interlaceprev and \interlacecurr as captured using our dual-exposure sensor (refer to the upper row). 
%The input frames are independently processed by a learned \makehdr network, so that the output sharp HDR frames \HDRsharp and blurry long exposure frames \longBlurry are obtained. 
%The input frames are independently processed by a learned \makehdr network so that the output sharp HDR images \HDRsharp are aligned with the short exposures, as well as blurry long exposure images \longBlurry, whose beginnings and ends are aligned with \HDRsharpStart and \HDRsharp are obtained. 
\mycfigure{overview}{
Overview of our HDR VFI pipeline. \emph{Upper row:} Two subsequent frames \interlaceprev and \interlacecurr, as captured using our dual-exposure sensor, are independently processed by a learned \makehdr network, so that the output sharp HDR frames \HDRsharp aligned with the end of the long exposure (the suffix {\suffixE} stands for the end) and blurry long exposure frames \longBlurry are obtained. 
Next, each frame is fed separately to a feature extractor to build feature pyramids. 
% \emph{Middle row:} At each pyramid scale $l$, we employ a learned \blurtoflow network
% to recover the intra-frame flow \flowInsideLevel given the features \HDRsharpFeat and \longBlurryFeat along with $\uparrow \flowInsideLevelprev$ upsampled from the previous scale. 
\emph{Middle row:} At each pyramid scale $l$, given the features \HDRsharpFeat and \longBlurryFeat along with $\uparrow \flowInsideLevelprev$ upsampled from the previous scale, the intra-frame flow \flowInsideLevel, the flow between the start (denoted with the suffix {\suffixS}) and end of the long exposure in each frame is recovered using a learned \blurtoflow network. 
We then find bidirectional flows \flowPrevCurrEndLevel and \flowCurrPrevEndLevel estimated between \sharpPrevEndLevel and \sharpCurrEndLevel (which are the sharp HDR frames \sharpPrevEnd and \sharpCurrEnd down-sampled by $2^l$) using the state-of-the-art flow estimation method Raft \cite{teed2020raft}. 
Next, given two estimated flows for each frame, we also derive additional flows of \flowPrevCurrStartLevel and \flowCurrPrevStartLevel. The motion flow triplets (\flowPrevCurrEndLevel, \flowPrevLevel, \flowPrevCurrStartLevel) as well as (\flowCurrPrevEndLevel, \flowCurrLevel, \flowCurrPrevStartLevel) are independently fed to a non-learnable \fitQuad module to calculate the forward flows \forwardprevLevel and \forwardcurrLevel that are parametrized using a quadratic motion model for a position $t$ (refer to the two bottom insets). Finally, using the module \blend, we fuse \sharpPrevEndLevel and \sharpCurrEndLevel with the forward flows \forwardprevLevel and \forwardcurrLevel and a soft occlusion map $\uparrow \alphaLevelPrev $
upsampled from the previous scale to reconstruct the intermediate frame \midFrameLevel at scale $l$. We repeat this procedure until we reach to the scale of the original input frames.
\emph{Bottom row:} A schematic presentation of all involved flows and their relation to the input and interpolated frames.}

\mysection{Method}{sec_method}
In this section, we propose a VFI method that reconstructs HDR frames in the continuous-time domain.
\refFig{overview} summarizes our processing pipeline, and the following paragraphs provide a more detailed description of its key components. 
Our method takes as input two subsequent video frames \interlaceprev and \interlacecurr that are captured using our dual-exposure sensor and produces a sharp HDR frame \midFrame for any position $t$ between \interlaceprev and \interlacecurr. Each captured frame \interlace, where with the suffix $i$ we denote any input frame, contains a pair of spatially interleaved short and long exposures and is processed by the \makehdr network to produce a sharp HDR frame \HDRsharp that is aligned with the end of the long exposure
(the suffix {\suffixE} stands for the end), and a blurry long exposure frame \longBlurry. 
Both frames are decomposed into their respective multi-resolution feature pyramids, and from this stage, the whole processing is performed at different scales, where as shown in the middle row in \refFig{overview}, information reconstructed at a lower-resolution scale $l+1$ contributes to the higher-resolution scale $l$.  
Here, for brevity, we omit the scale index $l$. The feature pyramids are fed to the \blurtoflow network to predict the flow \flowInside that extracts the flow between the start (denoted with the suffix {\suffixS}) and end of the long exposure.
% motion from the blur within the \longBlurry frame. 
Next, we compute the flows \flowPrevCurrEnd and \flowCurrPrevEnd between the sharp HDR frames \sharpPrevEnd and \sharpCurrEnd in both directions using an off-the-shelf flow estimation method such as Raft \cite{teed2020raft}. This way, we obtain the flows \flowPrev and \flowPrevCurrEnd that are aligned with \sharpPrevEnd, then we additionally derive the flow \flowPrevCurrStart, and employ all three flows to fit a quadratic motion model using a non-learnable \fitQuad module.
We repeat this process for the flows \flowCurr, \flowCurrPrevEnd, and \flowCurrPrevStart that are aligned with \sharpCurrEnd. 
Refer to the bottom row in \refFig{overview} for the depiction of the discussed flows.
Next, to warp the keyframes \sharpPrevEnd and \sharpCurrEnd to a novel temporal position $t$, we first find the forward flows \forwardprev and \forwardcurr and then compute the backward flows \backflowprev and \backflowcurr using 
% a differentiable flow reversal module (\reverse) 
differentiable flow reversal as introduced in \cite{xu2019quadratic}. Finally, using a multi-scale blending scheme \blend, we combine the warped images with a soft occlusion weight at different scales to synthesize the frame \midFrame.
We now provide more details on all the processing steps discussed here.

\myparagraph{HDR reconstruction: \makehdr} 
We acquire our input video using a dual-exposure sensor \cite{cmosis2020} that simultaneously captures a short and long exposure for each frame. In our setup, the exposure time for the long exposure is four times higher than the short exposure.
%(we experiment also with the exposure ratio of eight, and then, we state this explicitly). 
Each exposure is stored at odd and even columns in the sensor. As a result, both exposures are provided as half-resolution images, and they need to be up-sampled in the horizontal direction. Moreover, the short exposure exhibits strong noise in dark scene regions and requires denoising. On the other hand, the long exposure is less noisy, while it might contain considerable motion blur and requires deblurring. To do so, we employ the network design and the training strategy introduced in \cite{cogalan2022learning} to jointly deblur, denoise, and upsample our input frames \interlace to produce sharp, clean, and full-resolution short and long exposures. Both exposures are combined using a non-learnable technique, similar to \cite{debevec2008recovering}, to produce a sharp HDR frame \HDRsharp. We also extend the network output to produce an additional full-resolution blurry long exposure \longBlurry. 

\myparagraph{Motion from blur: \blurtoflow} 
As we discuss in \refSec{rw_motion_from_blur}, motion blur can potentially reveal information about the motion in the scene. We pursue this idea and propose the \blurtoflow network that derives the motion flow \flowInside that is associated with the blur pattern in the long exposure \longBlurry.  
The sensor design ensures that the short and long exposures are completed precisely at the same time point, and in our HDR reconstruction, the sharp frame \HDRsharp is aligned with the short exposure. Given \longBlurry and \HDRsharp provided in each frame, one can employ a standard motion estimation method to estimate the intra-frame flow. However, in our case, the two inputs are overlapping in time, and finding the correct correspondence of \HDRsharp in the long exposure \longBlurry is ambiguous. Therefore, an existing method such as PWC-Net \cite{pwcnet} cannot be adopted as is, so we apply the following modification to the PWC-Net architecture tailoring it to our inputs. In the original PWC-Net, the two nearby frames are fed to the same feature extractor to build the feature pyramids. Then, at each pyramid scale $l$, the feature of the second frame is warped to the position of the first frame using the upsampled flow, and a cost volume is created to compare the features of the first frame with the warped features from the second one. 
In our case, as the sharp HDR frame and long exposures are different in type, we process them with two independent feature extractors and create multi-scale features \HDRsharpFeat and \longBlurryFeat that correspond to the sharp HDR frame \HDRsharp and long exposure \longBlurry, respectively.
Then, at each scale $l$, the intra-frame flow \flowInsideLevel is estimated as follows:
\begin{align}
\flowInsideLevel = \blurtoflow( \HDRsharpFeat,\longBlurryFeat,\uparrow \flowInsideLevelprev)
\end{align}
\noindent
where \blurtoflow is a multi-layer CNN with DenseNet connections \cite{pwcnet,huang2017densely} and $\uparrow  \flowInsideLevelprev$ is the upsampled flow from the previous layer. Note at each scale, we do not need to warp the features of the sharp HDR frame, hence no cost volume must be computed. This process is repeated until a desired scale $l_0$ is reached.

\myparagraph{Quadratic motion model: \fitQuad}
%In this step, we explain our non-learnable quadratic motion modeling. 
We continue such multi-scale processing in our non-learnable quadratic motion modeling.
Given the intra-frame flows \flowPrevLevel and \flowCurrLevel
% (here the scale $l$ is omitted for simplicity) 
that are recovered  by \blurtoflow separately for each frame, we also find the inter-frame flows \flowPrevCurrEndLevel and \flowCurrPrevEndLevel between the HDR frames \sharpPrevEndLevel and \sharpCurrEndLevel (downscaled to a given scale $l$) using a state-of-the-art flow estimation method as proposed in \cite{teed2020raft}. While in practice, a quadratic motion model that is aligned with \sharpPrevEndLevel can be derived with only two flows (\flowPrevLevel and \flowPrevCurrEndLevel), we establish another possible flow, namely \flowPrevCurrStartLevel, which corresponds to the flow between \sharpPrevEndLevel and \sharpCurrStartLevel. It is computed as follows:
\begin{align}
\flowPrevCurrStartLevel = \flowPrevCurrEndLevel + \warp(\flowPrevCurrEndLevel,\flowCurrLevel)
\end{align}
\noindent
where \warp is a differentiable warping operator using bilinear sampling \cite{jaderberg2015spatial}. Here, the flow \flowCurrLevel is aligned with the frame \sharpCurrEndLevel; therefore, we need to warp \flowCurrLevel using the flow \flowPrevCurrEndLevel to become aligned with \sharpPrevEndLevel (refer to the bottom row in \refFig{overview}). Since the two flows are opposite in their directions, we sum up the flows instead of subtracting them.
Similarly, for the frame \sharpCurrEndLevel, we compute the additional flow \flowCurrPrevStartLevel as: 
\begin{align}
\flowCurrPrevStartLevel = \flowCurrPrevEndLevel + \warp(\flowCurrPrevEndLevel,\flowPrevLevel)
\end{align}
\noindent
Now, for warping \sharpPrevEndLevel to a novel time $t$, we derive a quadratic motion flow as:
\begin{align}
\forwardprevLevel = \quadmodelprev
\end{align}
\noindent 
where $ a_0$ and $v_0$ express the acceleration and velocity of a non-uniform motion, and they are derived from \flowPrevLevel, \flowPrevCurrEndLevel, and \flowPrevCurrStartLevel using the least square fit. Note that the derived model explains the non-uniform motion for the entire range of \sharpPrevStartLevel to \sharpCurrEndLevel. For a curvilinear motion, e.g. a rotatory motion, these parameters can be considered as the first two terms in the Taylor approximation of the curvilinear motion. Similarly, we can compute the flow \forwardcurrLevel:
\begin{align}
\forwardcurrLevel = \quadmodelcurr
\end{align}
\noindent
where the parameters $ a_1$ and $v_1$ are calculated from the triplet of flows \flowCurrLevel, \flowCurrPrevEndLevel, and \flowCurrPrevStartLevel using a least square fit. Existing VFI methods with non-uniform motion assumptions usually require more than two frames as the input. However, this enforces that the parameters of non-uniformity (acceleration and velocity) are fixed along multiple frames, which might not hold in practice. In contrast, our method only relies on two immediate frames, and as a result, we impose such constraints in a closer temporal range that allows us to model more complex non-uniform motion. Moreover, providing the additional flow \flowPrevCurrStartLevel not only allows us to approximate a higher order motion, e.g., a cubic motion model but also incorporates the motion flow information from the other frame to increase flow consistency between \forwardcurrLevel and \forwardprevLevel. 
In \refSec{abl_rw}, we ablate the effect of including \flowPrevCurrStartLevel and \flowCurrPrevStartLevel in our motion model. 
Since the time interval between \sharpCurrStartLevel and \sharpCurrEndLevel is shared when computing the motion model for the frame pairs \interlaceprev and \interlacecurr, and then \interlacecurr and \interlacecurrcurr, the temporal consistency is also preserved.

\myparagraph{Multiscale blending: \blend} In the last step, we introduce a multi-scale blending scheme to reconstruct the final interpolated image \midFrame. Specifically, at each scale $l$, given the forward flows \forwardprevLevel and \forwardcurrLevel, we compute the backward flows \backflowprevLevel and \backflowcurrLevel using the flow reversal %module  (\reverse) 
introduced in QVI \cite{xu2019quadratic}. We then warp the sharp HDR frames \sharpPrevEndLevel and \sharpCurrEndLevel to the novel position $t$ using the backward flows as:
\begin{align}
\warpPrevLevel =  \warp(\sharpPrevEndLevel,\backflowprevLevel) \textrm{ and }
\warpCurrLevel =  \warp(\sharpCurrEndLevel,\backflowcurrLevel)
\end{align}
where \sharpPrevEndLevel and \sharpCurrEndLevel are the input frames \sharpPrevEnd and \sharpPrevStart downsampled by $2^l$. Afterward, we predict the soft occlusion weight \alphaLevel that controls the contribution of input warped images \warpPrevLevel and \warpCurrLevel:
\begin{align}
\alphaLevel =  \blend(\warpPrevLevel,\warpCurrLevel,\backflowprevLevel,\backflowcurrLevel, \uparrow
\alphaLevelPrev)
\end{align}
\noindent
where \blend is a multilayer CNN and $\uparrow \alphaLevelPrev $ is the upsampled weight from the previous scale. Note the input flows \backflowprevLevel and \backflowcurrLevel aid the network in reasoning about the occlusion regions. Given the occlusion weight, the warped images are combined as follows:
\begin{align}
\midFrameLevel = \frac{(1-t) \alphaLevel \odot \warpPrevLevel + t (1-\alphaLevel) \odot \warpCurrLevel}{(1-t) \alphaLevel + t (1-\alphaLevel)} 
\end{align}
\noindent
where \midFrameLevel is the synthesized intermediate frame at scale $l$, as required in the loss computation (\refEq{e_HDR_loss}). The operator $\odot$ stands for per-pixel multiplication. Finally, at the finest scale $l_0$, the interpolated frame \midFrame is derived.

\myparagraph{Loss function}
Our loss function is composed of three components that are targeted to train the \makehdr, \blurtoflow, and \blend networks. 
First, the output of the \makehdr network is supervised with the ground truth \HDRsharpGT and \longBlurryGT (refer to \refSec{dataset_} on details of how we acquire the ground truth frames from high-framerate video datasets) using the reconstruction loss:
 \begin{align}
   \loss\mathrm{hdr} = \sum_{i=0,1} \lVert \HDRsharpGT - \HDRsharp \lVert_{1} +  \lVert \longBlurryGT - \longBlurry \lVert_{1}
 \end{align}
As the ground truth flow is not available, we employ a multiscale image loss to supervise the \blurtoflow network:
 \begin{align}
 \label{eq:loss_flow}
   \loss\mathrm{flow} = \sum_{i=0,1} \sum_{l=l_0}^{L}  \lVert \sharpPrevEndGTLevel - \warp(\sharpPrevStartGTLevel,\flowInsideLevel) \lVert_{1}
 \end{align}
 \noindent
where \sharpPrevEndGTLevel and \sharpPrevStartGTLevel are the ground truth frames \sharpPrevEndGT and \sharpPrevStartGT downsampled by $2^l$. At each scale $l$, we warp \sharpPrevStartGTLevel using the predicted flow \flowInsideLevel and compare with \sharpPrevEndGTLevel. Note that this loss component will try to align the warped image and the input frames for all regions in an image, including the occluded part. However, we argue this is not a significant issue because the intra-frame motion captured in the long exposure is relatively small compared to the inter-frame motion. Hence, we deal with small disoccluded areas within a frame, and the only degradation that can occur is over-smoothed flow at occlusion boundaries which can be resolved with a more sophisticated occlusion treatment.
Lastly, we supervise the output of the \blend network using the reconstruction loss at each scale:       
\begin{align}
\label{eq:e_HDR_loss}
   \loss\mathrm{synth} = \sum_{l=l_0}^{L} \lVert \midFrameGTLevel - \midFrameLevel \lVert_{1}
\end{align}

\noindent
where \midFrameGTLevel is the corresponding ground truth for interpolated frame \midFrameLevel at each scale $l$.
The final loss $ \loss\mathrm{total} $ is then computed as:
\begin{align}
   \loss\mathrm{total} = \loss\mathrm{hdr} +  \loss\mathrm{flow} + \loss\mathrm{synth}
\end{align}

It is worth mentioning that based on our observation, optimizing the network based solely on the final loss would create ambiguity as to whether the network should improve {\blurtoflow} or {\blend}  network to decrease the loss; therefore, intermediate supervision {(\refEq{loss_flow}} and {\refEq{e_HDR_loss})} is essential to train each component properly.

\myfigure{trajectory}{Trajectories of pixels (red dots) for 16 consecutive frames in a sample scene from four different datasets. The scenes in Adobe240 and GoPro datasets mostly have globally non-uniform motion due to non-uniform camera motion, while in datasets such as X4K1000FPS and SlowFlow, the scenes mostly contain locally non-uniform motion.}

\myfigure{histogram}{The histogram of measured non-uniform motions for different datasets, where the horizontal axis shows the normalized motion
error ($ \times 10^{-2} $) with respect to the linear motion fit, and the vertical axis denotes the probability of the observed frames given an error value. } 

% \begin{table}[]
%     \centering
%     \setlength{\tabcolsep}{4pt}
%     \caption{The statistics of scenes with uniform and non-uniform moving content for the Adobe240 \cite{su2017deep}, GoPro\cite{nah2017deep}, X4K1000FPS \cite{sim2021xvfi}, and SlowFlow \cite{janai2017slow} datasets. 
%     }
%     \vspace{0.1cm}
%     \resizebox{\columnwidth}{!}{
%     \begin{tabular}{l cccc}
%         \toprule
%         &
%         \multicolumn1c{\textbf{Adobe 240}}&
%         \multicolumn1c{\textbf{GoPro}} &
%         \multicolumn1c{\textbf{X4K1000FPS}} &
%         \multicolumn1c{\textbf{SlowFlow}} 
%         \\
        
%       \midrule

%         Uniform & \phantom{ }26 (20$\%$)  & \phantom{ }2 (\phantom{ }\phantom{ }6$\%$) & 108 ($98\%$) & 14 (35$\%$)\\
%         Non-uniform & 107 (80$\%$) & 31 ($94\%$) & \phantom{ }\phantom{  }2 (\phantom{ }\phantom{ }$2\%$) & 25 (65$\%$)\\
%     \bottomrule
%     \end{tabular}
%     \label{tbl:Nonlinearstats}}
%     %\includegraphics[width=\linewidth]{figures/MainAnalysis}
% \end{table}
% \raggedbottom

\mysection{Motion non-uniformity analysis}{sec_NonlinearityMetric}

In order to properly validate our proposed method, we must ensure that our dataset contains diverse examples of scene motion non-uniformity.  To this end, we analyze motion non-uniformity in some popular high-framerate video datasets, including Adobe240 \cite{su2017deep}, GoPro\cite{nah2017deep}, X4K1000FPS \cite{sim2021xvfi}, and SlowFlow \cite{janai2017slow}. 
Our procedure is as follows: For each pixel in a given frame, we use Raft \cite{teed2020raft} to track the corresponding pixels for $N$ consecutive frames. We choose $N=8$ for the Adobe240, GoPro, and SlowFlow datasets as they are captured with 240FPS, and eight frames represent the time gap between two consecutive frames in a 30FPS video, and we choose $N =33$ for X4K1000FPS containing 1000FPS videos. Note that in some cases, such tracking might fail due to occlusions and textureless regions. We find the occlusion regions by applying a forward-backward flow consistency check \cite{jonschkowski2020matters} between the first and last frames, and exclude them in our measurements. Likewise, as the estimated flow in the textureless regions is usually erroneous, we clip the flow to zero if its value is less than one pixel. \refFig{trajectory} shows the trajectories of pixels for four sample scenes that contain regions with non-uniform motion. In the next step, we find a linear model that, in the least square sense, fits the motion trajectory for each pixel. We then consider the mean square error with respect to such a linear fit, where higher errors indicate more motion non-uniformity. Note that for each pixel, the error value is normalized by the aggregated pixel displacement across the consecutive frames. Since the error is calculated for individual pixels, we measure the amount of motion non-uniformity in a frame by taking the 50th percentile of the calculated error over all pixels. We then repeat this procedure for non-overlapping sets of $N$ consecutive frames in each scene in each dataset. \refFig{histogram} shows the histogram of measured non-uniform motions for each dataset, where the horizontal axis denotes the error of the linear fit ($ \times 10^{-2} $) divided into eight discrete bins, and the vertical axis is the probability of observing the scene for a given error value. The Adobe240 and GoPro datasets feature significant percentages of non-uniform motion as they are  captured with a handheld camera.  Although large motions are present in the X4K1000FPS dataset, the camera moves along mostly linear trajectories. 

% In our experiments, we identify a scene as containing the non-uniform motion, when the average error is over $ 6.0\times 10^{-2} $ (the threshold value is chosen empirically based on our visual inspection of the pixel trajectories as illustrated in \refFig{trajectory}). Based on this threshold, we provide statistics of scenes with uniform and non-uniform motions for each dataset in \refTbl{Nonlinearstats}. 

\mysection{Implementation}{impl}
Our \makehdr network architecture follows  \cite{cogalan2022learning}. The network output is given in the Bayer domain, and we apply demosaicing using OpenCV \cite{opencv_library}, followed by a gamma correction to create the final short and long exposures in the sRGB format. The \blurtoflow network employs an architecture similar to the PWCNet \cite{pwcnet}, and also outputs the motion flow at a quarter
resolution and employs the context network for refining the flow. We then apply bilinear interpolation to obtain the half- and full-resolution flows. Our \blend network is implemented as a 12-layer conventional neural network with dilated convolutions and skip connections. During training, we use the patch size of $768 \times 768$; nevertheless, at the inference time, our convolutional network, as well as all non-learnable components, scale with resolution.

\mysection{Results}{Results_}
In this section, we first introduce the training and evaluation datasets.
Then we show quantitative and qualitative comparisons of our method with existing VFI methods. 
Finally, we provide ablation to justify our training set and different components of our method.

\mysubsection{Dataset}{dataset_}
As it is impossible to capture ground truth high-framerate HDR videos using our dual exposure sensor, and third-party high-framerate HDR videos are unavailable, we synthesize our training and evaluation datasets using existing LDR high-framerate videos. In our experiments, we take the scenes from X4K1000FPS \cite{sim2021xvfi} and SlowFlow \cite{janai2017slow} as our training datasets, and we consider Adobe240 \cite{su2017deep} and GoPro\cite{nah2017deep} as our evaluation datasets. Our training and testing video sequences are defined as follows: 
We take 16 consecutive frames in a high-framerate video, where the 1st and 4th frames are our sharp beginning and ending frames (\sharpPrevStart and \sharpPrevEnd). We sum up the four neighboring frames starting from 1 to 4 to simulate the long exposure \longBlurryPrev. We then skip 9 frames to simulate the camera readout gap. Similarly, we take the 13th and 16th frames as the \sharpCurrStart and \sharpCurrEnd and sum the frames from 13 to 16 to create the long exposure \longBlurryCurr. We consider frames 7 and 10 as the target frames for the reconstructions. Note that in our simulation of long exposures, we clip the aggregated pixel intensity if it exceeds the value of 255. In our simulation, we ignore each patch if more than 20\% of its content is already saturated in the original high-framerate video. In order to make our method robust to high blur and saturation, we perform data augmentation by creating different amounts of blur and different amounts of saturation. 
For our test set, we are interested in evaluating our method against the other methods for different ranges of non-uniformity; hence we split all scenes in the Adobe240 \cite{su2017deep} and GoPro\cite{nah2017deep} datasets into four different categories of Easy, Medium, Difficult, and Extreme based on the error magnitude of the linear fit derived in \refSec{sec_NonlinearityMetric}. Specifically, we divide the entire histogram range ($ 15 \times 10^{-2} $ here) into four equal segments (expressing our four motion non-uniformity categories), and we draw 125 sample frames both for the Adobe240 and GoPro datasets per each category. 

%This  each bin in \refFig{graph_test} reports the average reconstruction error for a given method over 200 samples per each category. 

% Specifically, we draw 25 sample frames from each of twenty bins in the histogram (\refFig{histogram}) both for the Adobe240 and GoPro datasets. Then, we divide the entire histogram range ($ 20 \times 10^{-2} $ here) into four equal segments (expressing our four motion non-uniformity categories), and we group all five bins inside each segment. Hence, we have overall 250 sample frames per each category.

\begin{table}[]
    \centering
    \setlength{\tabcolsep}{4pt}
    \caption{Quantitative comparison of our method with state-of-the-art VFI methods. The ABME and QVI methods are designed to handle non-uniform motions, while the XVFI and FILM methods rely on a linear motion assumption but can handle large motions. Methods are indicated with * when they are trained from scratch with our training set.}
    \vspace{0.2cm}
    % \columnwidth
    \resizebox{7cm}{!}{
    \begin{tabular}{lc cccccc}
        \toprule
        &
        \multicolumn2c{\textbf{Adobe240}}&
        \multicolumn2c{\textbf{GoPro}} & 
        \\
        \cmidrule(lr){2-3}
        \cmidrule(lr){4-5}
        \multicolumn1l{Methods}&
        \multicolumn1c{PSNR}&
      \multicolumn1c{ SSIM}&
        \multicolumn1c{PSNR}&
        \multicolumn1c{SSIM}&\\
      \midrule
    
    % \midrule
    ABME \cite{park2021asymmetric} &31.28 & 0.83& 30.98 & 0.82 \\
    QVI \cite{xu2019quadratic} & 31.30 & 0.86 & 30.80 & 0.84 \\
    QVI* \cite{xu2019quadratic} & 31.16 & 0.86 & 30.70 & 0.84\\
    \midrule
    XVFI \cite{sim2021xvfi} & 31.07 & 0.83 & 30.75 & 0.82\\
    XVFI* \cite{sim2021xvfi} & 30.66 & 0.83 & 30.41 & 0.82\\
    FILM \cite{reda2022film} & 31.11 & 0.83 &  30.75 & 0.82\\
    FILM* \cite{reda2022film} & 31.04 & 0.83 & 30.74 & 0.82\\
    \midrule
    Ours  & \textbf{34.82} & \textbf{0.93} & \textbf{35.01} & \textbf{0.92} \\

    \bottomrule
    \end{tabular}}
    \label{tbl:numbers}
\end{table}
\raggedbottom

% \begin{table}[]
%     \centering
%     \setlength{\tabcolsep}{4pt}
%     \caption{Quantitative comparison of our method with state-of-the-art VFI methods. The ABME and QVI methods are designed to handle non-uniform motions, while the XVFI and FILM methods rely on a linear motion assumption but can handle large motions.}
%     \vspace{0.2cm}
%     \resizebox{4.5cm}{!}{
%     \begin{tabular}{lc cccc}
%         \toprule
%         \multicolumn1l{Methods}&
%         \multicolumn1c{PSNR}&
%       \multicolumn1c{ SSIM}\\
%       \midrule

%     % BIN \cite{shen2020blurry} & & & &  \\
%     % UTI-VFI  \cite{zhang2020video} & & & &  \\
    
%     % \midrule
%     ABME \cite{park2021asymmetric} & 31.13 & 0.83\\
%     QVI \cite{xu2019quadratic} &31.05 & 0.84 \\
    
%     \midrule
%     XVFI \cite{sim2021xvfi} & 30.91 & 0.82 \\
%     FILM \cite{reda2022film} & 30.93 & 0.82& \\
    
%     \midrule
%     Ours  & \textbf{34.37} & \textbf{0.91}  \\
%     % Ours+denoise & 34.12 & 0.93 & 32.98 & 0.90\\
%     \bottomrule
%     \end{tabular}}
%     \label{tbl:numbers}
%     %\includegraphics[width=\linewidth]{figures/MainAnalysis}
% \end{table}
% \raggedbottom

\myfigure{graph_test}{Quantitative comparison of our method with state-of-the-art VFI methods for four different motion non-uniformity categories 
(refer to \refSec{dataset_}). Each bin reports the average reconstruction error for a given method over 250 sample frames per category.}

%The ABME and QVI methods are designed to handle non-uniform motions, while the XVFI and FILM methods rely on a linear motion assumption but can handle large motions.

\mycfigure{flow_vis_real}{Visualization of flow maps reconstructed by our {\blurtoflow} network. Otherwise, the figure layout follows the one in \refFig{teaser}.}

\mycfigure{results_synthetic}{Visual comparisons of our method with the state-of-the-art VFI methods using the synthetic dataset described in \refSec{dataset_}. 
For each of the three scenes, the first row of insets shows the performance of respective VFI methods, while the second row presents the corresponding per-pixel error maps between the interpolated results and the ground truth. The PSNR/SSIM values written below each error map are computed for each inset rather than the entire image. In the upper scene taken from the X4K1000FPS test set, the wheel moves in a non-linear trajectory, and the existing VFI methods struggle to position the wheel correctly for the interpolated frames, while our method leads to a good alignment with the ground truth. In the middle scene taken from the GoPro dataset, the camera is moving with an extremely non-uniform motion as shown in \refFig{trajectory}. While the existing VFI methods produce visually plausible results, they are not correctly aligned with the ground truth as the error map reveals. The bottom scene, taken again from the X4K1000FPS test set, contains a combination of camera and object movements. In this case, the existing VFI methods fail to properly handle the occlusion boundaries.}

% \begin{table}[]
%     \centering
%     \setlength{\tabcolsep}{4pt}
%     \caption{Quantitative comparison of our method with state-of-the-art VFI methods. The ABME and QVI methods are designed to handle non-uniform motions, while the XVFI and FILM methods rely on a linear motion assumption but can handle large motions.}
%     \vspace{0.2cm}
%     \resizebox{\columnwidth}{!}{
%     \begin{tabular}{lc cccccc}
%         \toprule
%         &
%         \multicolumn2c{\textbf{Uniform}}&
%         \multicolumn2c{\textbf{Non-uniform}} & 
%         \\
%         \cmidrule(lr){2-3}
%         \cmidrule(lr){4-5}
%         \multicolumn1l{Methods}&
%         \multicolumn1c{PSNR}&
%       \multicolumn1c{ SSIM}&
%         \multicolumn1c{PSNR}&
%         \multicolumn1c{SSIM}&\\
%       \midrule

%     % BIN \cite{shen2020blurry} & & & &  \\
%     % UTI-VFI  \cite{zhang2020video} & & & &  \\
    
%     % \midrule
%     ABME \cite{park2021asymmetric} &33.16 & 0.88& 31.04& 0.83 \\
%     QVI \cite{xu2019quadratic} &32.10 & 0.89 & 31.32 & 0.85 \\
    
%     \midrule
%     XVFI \cite{sim2021xvfi} & 32.56 & 0.87 & 30.88 & 0.82\\
%     FILM \cite{reda2022film} & 32.89& 0.87& 30.90& 0.82\\
    
%     \midrule
%     Ours  & \textbf{35.97} & \textbf{0.94} & \textbf{35.57} & \textbf{0.93} \\
%     % Ours+denoise & 34.12 & 0.93 & 32.98 & 0.90\\
%     \bottomrule
%     \end{tabular}}
%     \label{tbl:numbers}
%     %\includegraphics[width=\linewidth]{figures/MainAnalysis}
% \end{table}
% \raggedbottom

\mysubsection{Quantitative comparison}{Quantitative_}
We compare our proposed method with state-of-the-art sharp VFI methods (refer to \refSec{rw_SharpInterp}): FILM \cite{reda2022film} and XVFI \cite{sim2021xvfi} which rely on a uniform motion assumption, and QVI \cite{xu2019quadratic}, and ABME \cite{park2021asymmetric} which explicitly support the non-uniform motion.
%FILM, XVFI, and ABME require just two neighboring frames as input, while the method QVI needs to process four consecutive frames. 
QVI employs four consecutive frames as the input, and FILM and XVFI require just two frames.
While ABME also uses only two frames as input, it relaxes the uniform motion constraint by first estimating symmetric bilateral motion fields and then refining them to become asymmetric. 
% Since these VFI methods expect post-processed sRGB images as input, and moreover, we are interested in evaluating our interpolation component, we feed the sharp HDR output to the sharp frame interpolation methods.
As the LDR (sRGB) images in the high-framerate dataset are used to synthesize our training and evaluation set, we can directly feed them as input to the VFI methods. For our method, though, we feed them along with the simulated long exposure as described in \refSec{dataset_}.
Note that we are unable to compare with the blurry VFI methods (refer to \refSec{rw_DeblurInterp}), as they require well-exposed blurry input frames (effectively, blurry HDR frames) while our long exposure typically contains a considerable amount of saturation that poorly handled by these methods. 
\refTbl{numbers} summarizes our comparisons with the VFI methods (used with their pre-trained weights) for each of our test datasets (Adobe240 and GoPro) separately as specified in \refSec{dataset_}. Note that XVFI uses almost the same training set as ours while applying extra data augmentation, and a method such as FILM carefully prepared their dataset to include all the possible motion ranges, with a much larger training data size than we consider. Nevertheless, for a fair comparison, we have re-trained XVFI, FILM, and QVI using our training set (indicated with * in {\refTbl{numbers}}) and observed a lower performance. Unfortunately, the training code for ABME is not publicly available. Moreover, \refFig{graph_test} provides a deeper insight into each method performance when we aggregate those datasets and split them into four different categories with respect to motion complexity (\refSec{sec_NonlinearityMetric}).
Overall for more uniform motion, the competing VFI methods perform similarly, while clear advantages of the QVI method can be seen for more complex motion. 
In all cases, our method outperforms the existing VFI methods by a large margin. It is also more stable in the interpolation quality for higher motion non-uniformity. We hypothesize that this stability could be attributed to our quadratic motion fitting part, which has no learnable parameters and only relies on the accuracy of flows, which might drop off slightly at higher non-uniform motion.
Other VFI solutions that mostly learn how to handle non-uniform motion might impose higher requirements on the training set.

% In an informal experiment, we confirmed that removing highly non-linear motion from our training dataset does not affect the performance of our technique.
\mycfigure{real_results}{The visual comparisons of our interpolation results for three scenes captured using our camera with a dual-exposure sensor.
Our method is able to correctly interpolate the frames in the scenes with the challenging cases of a rolling disc (the upper scene), a rotary camera motion (the middle scene), and a moving object behind a refractive object (the bottom scene). }
     
\mysubsection{Qualitative comparison}{Qualitative_}

We first visualize the examples of HDR scenes captured in daylight and dark conditions in \refFig{teaser} and \refFig{flow_vis_real}.
The flow map reconstructed by our {\blurtoflow} module in \refFig{flow_vis_real}, as well as the motion blur magnitude in the long exposures indicate the complexity of motion. 
%where the magnitude of the motion blur in the long exposure can show the amount of motion. 
In the accompanying videos, we demonstrate that competing VFI methods struggle with the scene in \refFig{teaser}, while our method benefits from additional information that is encoded in the motion blur pattern to improve the interpolation quality. 
%Additionally, \refFig{flow_vis_real} provides the flow map reconstructed by our {\blurtoflow} module. 
We then provide visual comparisons with the state-of-the-art VFI methods for three synthesized scenes with ground truth in \refFig{results_synthetic}. Moreover, we compare to other methods in \refFig{real_results} using the captured sequences. All the capturing processes were done with our Axiom-beta camera with a CMOSIS CMV12000 sensor \cite{cmosis2020}. In both setups, we use the exposure ratio of 4 between the short and long exposures. Since the frames captured using our camera cannot be fed directly to the other VFI methods, we first reconstruct the sharp HDR images {\sharpPrevEnd} and {\sharpCurrEnd} using our {\makehdr} network. They are then tonemapped using Reinhard-Global 2002 \cite{reinhard2002photographic} and gamma-corrected, and are fed to the LDR VFI methods. The upper scene in \refFig{real_results} shows an example of a rolling disc in which the existing VFI methods, even the ones designed to deal with non-uniform motion such as ABME and QVI, fail to properly interpolate an intermediate frame due to non-uniform motion caused by the rotatory motion of the disc. In the next examples, we captured a crystal ball while the camera is rapidly rotating (the middle scene) or an object is moving behind the crystal ball (the bottom scene). We can  observe that in these challenging examples where even a uniform motion in the scene might appear non-uniform in the refracted image, other methods struggle to correctly reconstruct an in-between frame. In all cases, we can see our method faithfully reconstruct the in-between frames even in difficult conditions where there are reflections on the crystal ball (the middle and bottom scenes). Please refer to our supplementary video for the temporal consistency of our method.

\myfigure{failure_case}{ Our interpolation failure example in a case where the moving content is highly saturated in the long exposure.}

% \begin{table}[ht]
%     \centering
%     \setlength{\tabcolsep}{4pt}
%     \caption{The effect of including the scenes with non-uniform motion in our training dataset. 
%     }
%     \vspace{0.2cm}
%     \resizebox{\columnwidth}{!}{
%     \begin{tabular}{lc cccccc}
%         \toprule
%         &
%         \multicolumn2c{\textbf{Uniform}}&
%         \multicolumn2c{\textbf{Non-uniform}} & 
%         \\
%         \cmidrule(lr){2-3}
%         \cmidrule(lr){4-5}
%         \multicolumn1l{Methods}&
%         \multicolumn1c{PSNR}&
%         \multicolumn1c{SSIM}&
%         \multicolumn1c{PSNR}&
%       \multicolumn1c{ SSIM}\\
%       \midrule
%         Ours-uniform  & 34.86 & 0.93 & 34.42 & 0.91 \\
%         Ours-non-uniform  & 34.42 & 0.93 & 34.10 & 0.91\\
%     \bottomrule
%     \end{tabular}
%     \label{tbl:trainingset}}
%     %\includegraphics[width=\linewidth]{figures/MainAnalysis}
% \end{table}
% \raggedbottom

\mycfigure{fig_ablation}{Ablation results. The figure layout is similar to \refFig{results_synthetic}. Refer to \refSec{abl_rw} for more details on each ablation scenario. }

\mysubsection{Ablation study}{abl_rw}

We perform a series of ablations to show the contributions of each key component in our proposed method and to analyze the alternative solutions. 
We summarize the obtained results in \refFig{fig_ablation} and \refTbl{ablations}, where each ablation component we denote with a unique label that is also included in the related paragraph title.

%\myparagraph{Impact of training with uniform motion data: OnlyUniform} We perform an experiment to show the effect of including scenes with non-uniform motion in our training dataset. We train our method using only the scenes with error of linear fit close to zero according to \refSec{sec_NonlinearityMetric}. As can be seen in \refTbl{ablations} our results show no significant difference in the performance of our method when trained using scenes with non-uniform motion. This is due to the fact that the quadratic motion fitting part of our method does not have any learnable parameters that could benefit from non-uniform motion samples in the training set.
  
\myparagraph{Impact of \blurtoflow network: NoBlur2Flow} We analyze the contribution of the \blurtoflow network where we attempt to reconstruct the intermediate frames using only the backward and forward flows between the sharp HDR frames \sharpPrevEnd and \sharpCurrEnd using Raft \cite{teed2020raft}. 
% In this case, 
This experiment suggests the version of our method that makes the linear assumption, in which we linearly split the flow at any position $t$ between the frames; however, this leads to large positional errors in the interpolated content, as seen in \refFig{fig_ablation}.
Our results clearly indicate the effectiveness of including the \blurtoflow network in our pipeline (\refTbl{ablations}).

\myparagraph{Impact of sharp HDR frame: NoSharp} We investigate the effect of including the sharp HDR frame \HDRsharp, along with the long blurry exposure \longBlurry, on the accuracy of motion from blur derivation. 
To do so, we consider \longBlurry as the only input to the \blurtoflow network and exclude \HDRsharp (note that \HDRsharp is still available for other components in our pipeline).
As it can be seen in \refFig{fig_ablation} the availability of \HDRsharp reduces geometric image distortions and \HDRsharp compensates for the lack of information for saturated pixels that are inherent for \longBlurry in our setup with a dual-exposure sensor. 
Following this observation, we expect that replacing our \blurtoflow network with a solution, where the intra-frame flow is extracted solely based on \longBlurry \cite{zhang2020video} should lead to a similar outcome as this ablation.

\myparagraph{Quadratic model with temporal flows: TemporalFlows}Considering more than two consecutive frames involves a larger time span; as a result, fine-grained motion cannot be properly handled. We have made such observations when comparing our method with a method like QVI, which uses four frames to compute the quadratic model. Nonetheless, to highlight the advantage of the intra-flow {\flowInside} estimated from the {\blurtoflow} module, we conduct an ablation where we fit the quadratic motion using the temporal flows extracted from four consecutive HDR frames (similar to QVI); however, we observed a lower performance than ours with two frames, while it still has a better performance compared to QVI.

\myparagraph{Alternative approach to \blurtoflow network: SharpStart} Instead of directly recovering the motion flow from the blur, we employ a 12-layer conventional neural network with dilated convolutions to predict the sharp frame \HDRsharpStart aligned with the beginning of the frame, then use the Raft \cite{teed2020raft} to estimate the intra-frame flow \flowInside between the \HDRsharp and predicted \HDRsharpStart. 
This ablation demonstrates that the particular method of deriving the intra-frame flow from motion is less important, under the condition that sharp, saturation-free reference \HDRsharp is available.  Still, our proposed method leads to slight quality improvement.

\myparagraph{Quadratic vs. cubic motion model: CubicModel} Since our method provides three estimated flows in each frame, we are able to approximate a higher-order motion, e.g., cubic. Hence, we perform an ablation where we replace the quadratic motion model derived in \refSec{sec_method} with a cubic model. 
Overall the obtained results are comparable in terms of the SSIM prediction, but the quadratic model is slightly better in terms of PSNR and visual results (\refFig{fig_ablation}). A key difference is that while the cubic model involves a closed-form solution, we derive the quadratic model in a least-squares fashion that allows for the correction of slight errors in the derived flows.

\myparagraph{Two vs. three flows: TwoFlows} To see the effect of including the additional flows \flowCurrPrevStart and \flowPrevCurrStart in the derivation of our quadratic motion model, we exclude them from the input to the \fitQuad module.  
The obtained results (\refTbl{ablations}) indicate that including an independent estimate of the third flow contributes toward correcting for potential inconsistencies in the other two flows. For example, in \refFig{fig_ablation}, ghosting artifacts along higher contrast edges are clearly visible when only two flows are employed.

%\myparagraph{Single- vs. multi-scale blending: SingleBlending} In order to show the effect of our multi-scale blending scheme, we apply a single-stage blending where the \blend network takes \warpPrev and \warpCurr along with \forwardcurr and \forwardprev at the original input resolution and directly outputs the soft occlusion mask.     

\begin{table}[]
    \centering
    \setlength{\tabcolsep}{3pt}
    \caption{The ablation results indicate the performance of alternative solutions for major design choices in our proposed method. Refer to \refSec{abl_rw} where we provide more details on each ablation. % and we refer to them using the same labels as those used in the table. 
    }
    \vspace{0.2cm}
    \resizebox{4.0cm}{!}{
    \begin{tabular}{lc cccc}
        \toprule
        &
        \multicolumn1c{PSNR}&
       \multicolumn1c{ SSIM}\\
       \midrule

    % BIN \cite{shen2020blurry} & & & &  \\
    % UTI-VFI  \cite{zhang2020video} & & & &  \\
    
    % \midrule
 %   OnlyUniform &  & \\
    NoBlur2Flow & 30.97 & 0.82 \\
    NoSharp & 30.28 & 0.82 \\
    TemporalFlows & 31.93 & 0.85 \\
    SharpStart & 34.35 & 0.91 \\
    CubicModel & 33.84 & 0.92 \\
    TwoFlows & 34.00 & 0.90 \\
  %  SingleBlending & 34.89 & 0.92\\
    
    \midrule
    Ours  & \textbf{34.92} & \textbf{0.93}  \\
    \bottomrule
    \end{tabular}}
   \label{tbl:ablations}
\end{table}
\raggedbottom

%\begin{table}[ht]
%    \centering
%    \setlength{\tabcolsep}{4pt}
%    \caption{The effect of \blurtoflow network in our pipeline. 
%    }
%    \vspace{0.1cm}
%    \resizebox{\columnwidth}{!}{
%    \begin{tabular}{lc cccccc}
%        \toprule
%        &
%        \multicolumn2c{\textbf{Uniform}}&
%        \multicolumn2c{\textbf{Non-uniform}} & 
%        \\
%        \cmidrule(lr){2-3}
%        \cmidrule(lr){4-5}
%        \multicolumn1l{Methods}&
%        \multicolumn1c{PSNR}&
%        \multicolumn1c{SSIM}&
%        \multicolumn1c{PSNR}&
%       \multicolumn1c{ SSIM}\\
%       \midrule
%        Ours-w/o blur2flow  & 31.51 & 0.86 & 30.10 & 0.81 & &\\
%        Ours-full  & \textbf{35.97} & \textbf{0.94} & \textbf{35.57} & \textbf{0.93}\\
%    \bottomrule
%    \end{tabular}
%    \label{tbl:blurflow}}
%    \includegraphics[width=\linewidth]{figures/MainAnalysis}
%\end{table}
%\raggedbottom

% \begin{itemize}
%   \itemsep0em 
%   \item showing the effect of including the scenes with non-uniform motion in our training dataset (the test does always have non-uniform motion)
%   \item showing the contribution of each flow (blur2flow and large flow and combined flow) providing the table and visual
%   \item in the term of capturing the non-uniformity of the motion we need to justify that flowInside provide more reliable information compared to using three temporal  
% \end{itemize}

% \mysubsection{Effect of exposure on frame interpolation}{Effect_}
%  We want to confirm it visually that well-exposed scene is better for frame interpolation compared single exposure.
%  for example moving hand in front of window 

\mysubsection{Limitations and future work}{Limitation_}
Saturation is inevitable in long exposure for bright scene regions. 
% If moving content is present in such regions, 
In case of a local motion blur that is fully covered with saturation, our flow prediction using the \blurtoflow network becomes less accurate. \refFig{failure_case} shows an example of this case where we synthetically increase the saturation in the long exposure for the wheel example shown in \refFig{results_synthetic}, and our method fails to correctly reconstruct the intermediate frame. However, in case of a local motion blur with partial saturation or a global camera motion, even with fully saturated regions, as shown in {\refFig{results_synthetic} and {\refFig{fig_ablation}}, our method can recover the flow by propagating the flow information from the unsaturated regions.}

The dynamic range that we can reconstruct is limited by the exposure ratio of four that we assume in this work. For larger ratios, the accuracy of HDR frame reconstruction by the \makehdr network might be reduced \cite{cogalan2022learning}, which could adversely affect the accuracy of HDR video  interpolation. Moreover, when capturing an HDR scene, we adjust the lowest exposure time in such a way that the long exposure is not very saturated so that there is enough valuable blurry information. This procedure is currently done manually; an automatic selection of the optimal exposure time is an interesting future work direction that could lead to further performance improvements. We also relegate as future work porting our technique to other multi-exposure sensors that are used in modern smartphones \cite{quadBayerphone}, such as Sony's Quad Bayer \cite{quadBayer} and Samsung's Tetracell/Nonacell \cite{samsungisocell-gn1} sensors.
Such sensors should enable further improvements in the VFI quality via a more uniform layout of pixels with varying exposures. It would also be interesting to experiment with more than two exposures, as supported by such sensors. 

Lastly, investigating optical blur and finding ways to remove it along with motion blur could be an interesting, but challenging, future direction. The current state-of-the-art image restoration methods \cite{zamir2022restormer,wang2022uformer} still treat them as two separate tasks due to the difficulties in removing the coexisting blur. However, we believe our employed sensor design can significantly facilitate disentangling the motion blur that changes with exposure from the optical blur that remains constant between exposures.

\mysection{Conclusion}{Conclusion_}
In this work, we presented a method for high-dynamic-range video frame interpolation using dual-exposure sensors.
Our method outperforms the existing VFI methods both in terms of quantitative metrics as well as visual results for the challenging scenes containing non-uniform motions.
In particular, we achieve high-precision alignment of scene motion with the ground truth, where other methods clearly fail, although they may produce visually plausible results.
Our method can handle complex motion with consistently high performance as it depends little on explicitly training this reconstruction aspect.
Instead, we capitalize on the increased temporal sampling rate due to motion reconstruction from blur information.
Also, our method is less dependent on scene lighting conditions, whereas other methods designed for single-exposure sensors may suffer from image saturation in bright regions or excessive noise in dark conditions.

%\clearpage
{\small
\bibliographystyle{eg-alpha}
\bibliography{egbib}
}

\end{document}